\documentclass[letterpaper, 10 pt, conference]{ieeeconf}  

\IEEEoverridecommandlockouts 

\usepackage{amsmath,amssymb,amsfonts}
\usepackage{graphicx}
\usepackage{textcomp}
\usepackage{xcolor}
\usepackage{float}
\usepackage{amsmath}
\usepackage{booktabs}
\usepackage{algorithm2e}
\LinesNumbered

\usepackage{subfigure}
\usepackage{booktabs,multirow,tabularx,makecell}
\usepackage[table]{xcolor}
\usepackage{textcase}
\usepackage[font=footnotesize]{caption} 

\usepackage{censor}

\newcommand{\squeezeWords}{\looseness=-1}

\newcommand{\pushLines}{\pagebreak}

\newcolumntype{Y}{>{\centering\arraybackslash}X}

\DeclareMathOperator*{\argmin}{arg\,min}

\RestyleAlgo{ruled}
\SetAlgoInsideSkip{2pt}
\SetKwInput{KwIn}{Given}
\SetKw{KwInit}{Initialize}
\SetNlSty{}{}{}
\SetAlgoNlRelativeSize{-1}
\SetInd{0.6em}{1.1em}

\newcommand{\Ball}[2]{B(#1,#2)}
\newcommand{\Xobs}[1]{X_{\mathrm{obs}}(#1)}
\newcommand{\Xsense}[1]{X_{\mathrm{sensed},#1}}
\newcommand{\Xfree}[1]{X_{\mathrm{free},#1}}

\makeatletter
\let\NAT@parse\undefined
\makeatother
\usepackage[numbers,sort&compress]{natbib} 

\usepackage[pdftex,colorlinks=true,linkcolor=black,citecolor=black,urlcolor=black]{hyperref}
\usepackage{hypernat} 

\title{\LARGE \bf
Revisiting Replanning from Scratch: Real-Time Incremental Planning with Fast Almost-Surely Asymptotically Optimal Planners
}

\author{Mitchell E. C. Sabbadini$^{1}$, Andrew H. Liu$^{2}$, Joseph Ruan$^{2}$, Tyler S. Wilson$^{1}$,\\ Zachary Kingston$^{2}$, and Jonathan D. Gammell$^{1}$
\thanks{$^{1}$ Estimation, Search, and Planning (ESP) Research Group, Queen’s University, Kingston ON, Canada. {\tt\small 
\{20ms116,18tsw1,gammell\}@queensu.ca.
}} 
\thanks{
$^{2}$ CoMMA Lab, Department of Computer Science, Purdue University, West Lafayette IN, USA \tt\small \{liu3458, ruan44, zkingston\} @purdue.edu.%
} 
} 

\begin{document}

\maketitle
\thispagestyle{empty}
\pagestyle{empty}

\bibliographystyle{IEEEtran}


\begin{abstract}

Robots operating in changing environments either predict obstacle changes and/or plan quickly enough to react to them. 
Predictive approaches require a strong prior about the position and motion of obstacles. 
Reactive approaches require no assumptions about their environment but must replan quickly and find high-quality paths to navigate effectively.

Reactive approaches often reuse information between queries to reduce planning cost. 
These techniques are conceptually sound but updating dense planning graphs when information changes can be computationally prohibitive. 
It can also require significant effort to detect the changes in some applications.

This paper revisits the long-held assumption that reactive replanning requires updating existing plans. 
It shows that the incremental planning problem can alternatively be solved more efficiently as a series of independent problems using fast almost-surely asymptotically optimal (ASAO) planning algorithms. 
These ASAO algorithms quickly find an initial solution and converge towards an optimal solution which allows them to find consistent global plans in the presence of changing obstacles without requiring explicit plan reuse. 
This is demonstrated with simulated experiments where Effort Informed Trees (EIT*) finds shorter median solution paths than the tested reactive planning algorithms and is further validated using Asymptotically Optimal RRT-Connect (AORRTC) on a real-world planning problem on a robot arm.

\end{abstract}

\section{Introduction}

A key application of motion planning algorithms in robotics is to find paths in environments with unknown or dynamic obstacles. 
Most planners use some combination of predictive or reactive approaches to find feasible or optimal paths quickly and efficiently \cite{22-grothe-ICRA}. 
Predictive approaches attempt to predict the movement of obstacles in the environment and generate plans around these predictions. 
Reactive approaches make exact plans using the known information and quickly modify them when changes are detected.

Predictive planners leverage \textit{a priori} knowledge of their likely environments to improve their planning efficiency and path quality. 
When this includes the trajectories of dynamic obstacles, the planner can find plans that guarantee it does not collide with obstacles \cite{22-grothe-ICRA}. 
Planners may also inform their paths by avoiding sections of the environment that are likely to contain an obstacle \cite{luders2010chance} or ensuring their paths are valid within assumed obstacle parameters \cite{fiorini1998velocity}.

\begin{figure}[!t]
  \centering
  \includegraphics[width=\linewidth]{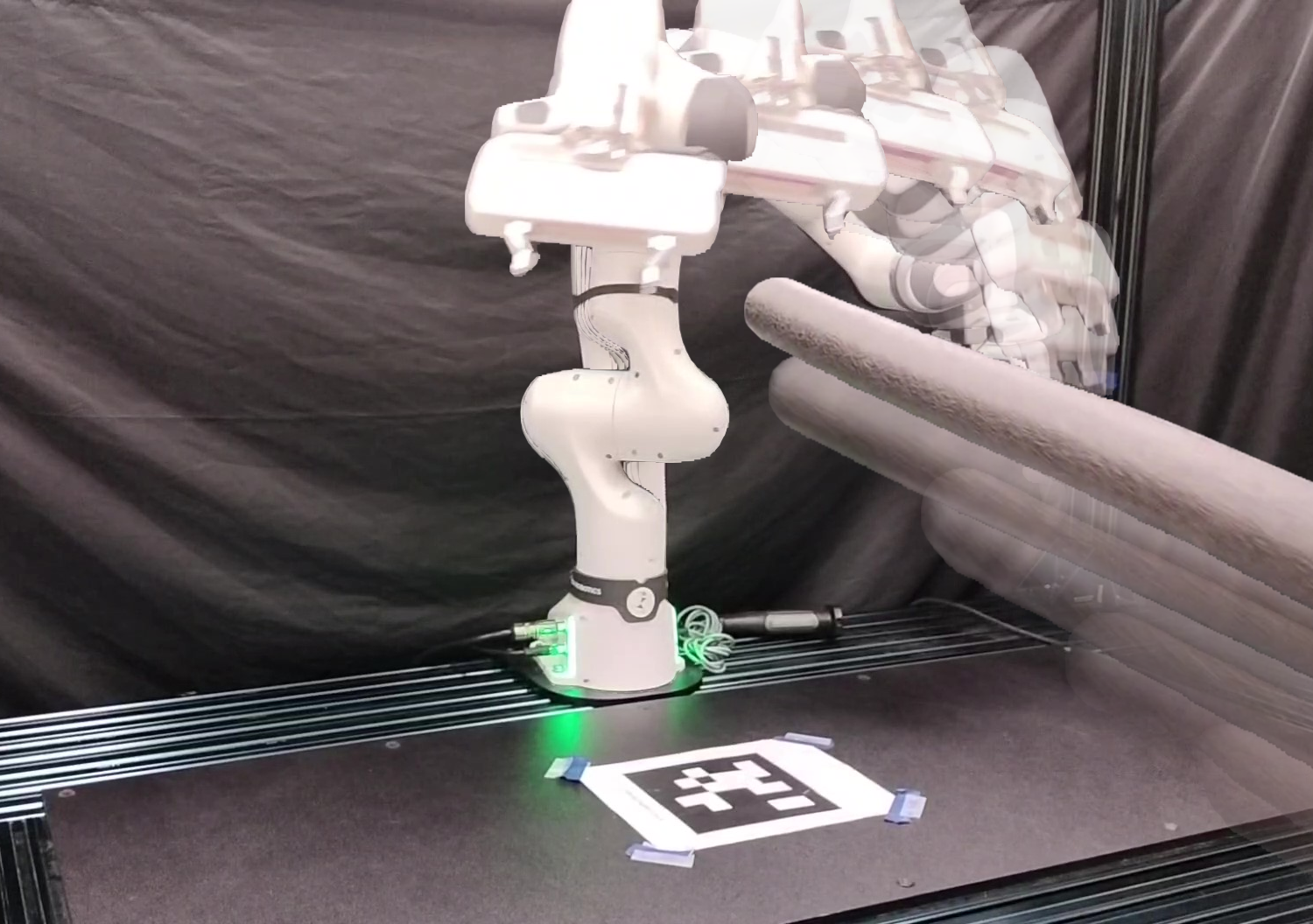}
  \setlength{\belowcaptionskip}{-14pt}
  \caption{
  A Franka Research 3 arm replanning around a dynamic obstacle by independently solving new planning problems each time the obstacle information changes. 
  The VAMP \cite{thomason_motions_microseconds_2024} implementation of AORRTC \cite{wilson2025aorrtc} plans fast enough in a CAPT \cite{Ramsey-RSS-24} obstacle map that it is able to replan without stopping the arm.
  }
  \label{fig:stacked_experiments}
\end{figure}

\begin{figure*}[t]
  \centering
  \subfigure[EIT*]{\includegraphics[scale=0.25]{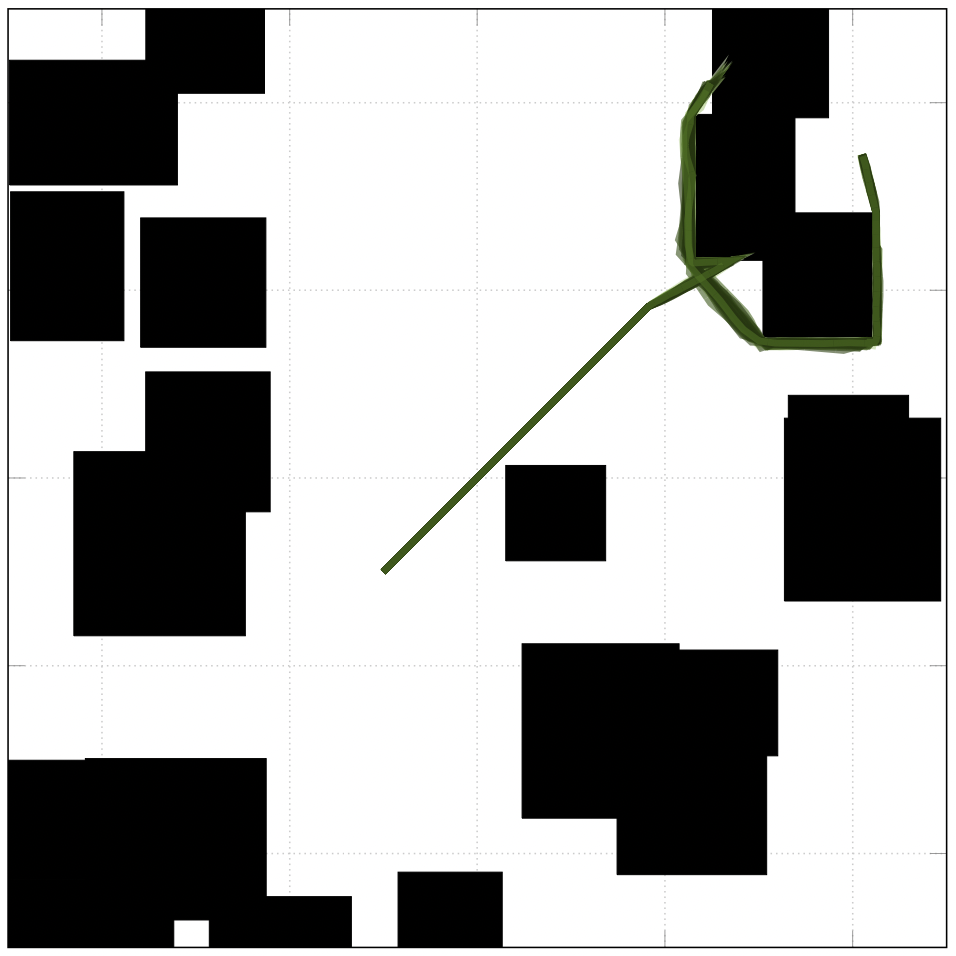}}
  \subfigure[RRT*]{\includegraphics[scale=0.25]{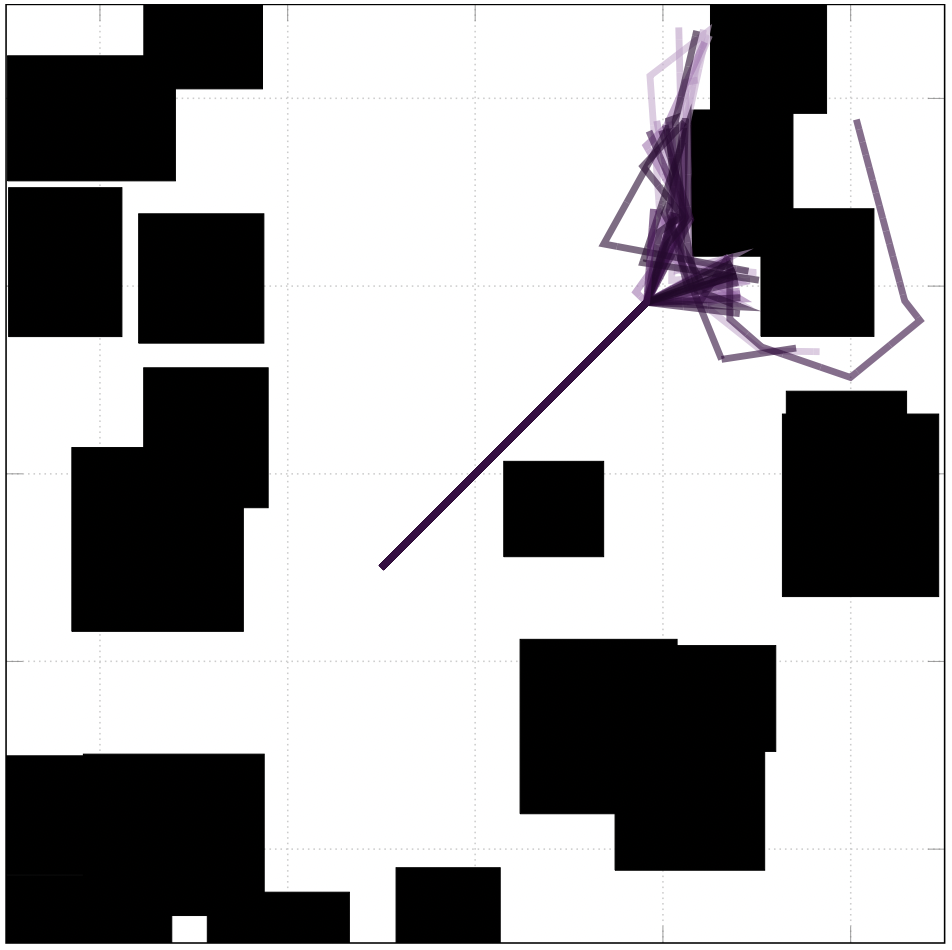}}
  \subfigure[RRT-Connect]{\includegraphics[scale=0.25]{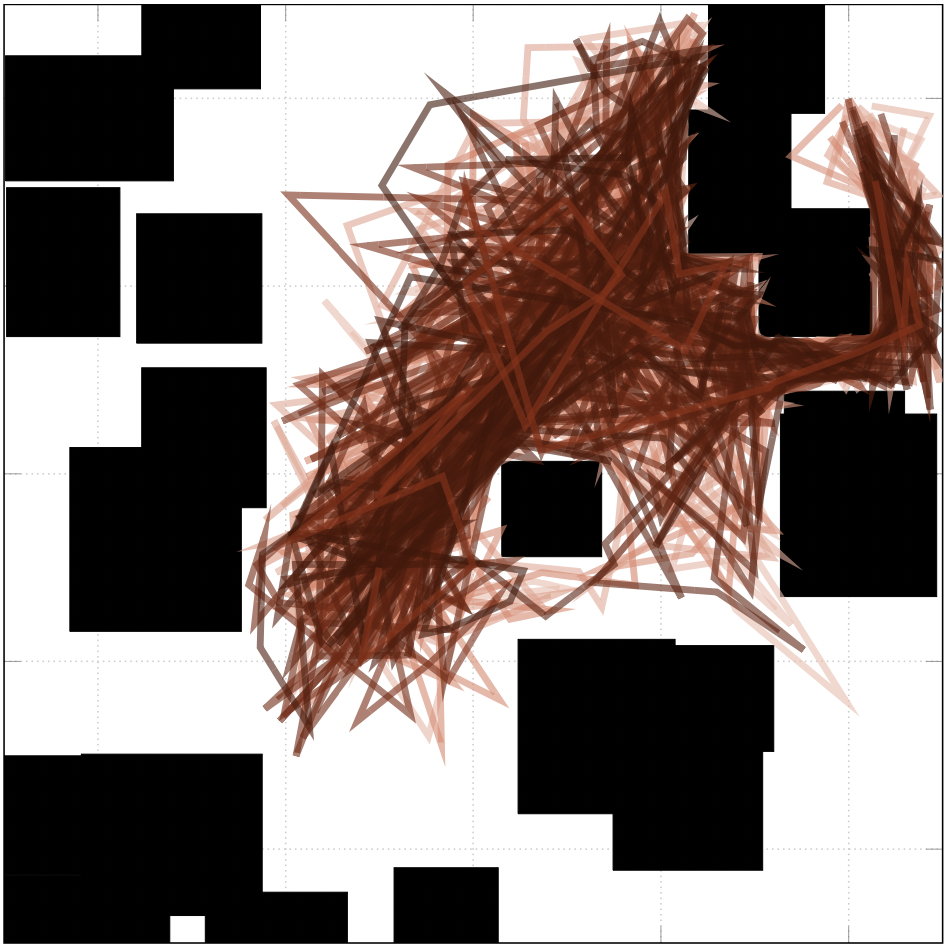}}
  \subfigure[RRT\textsuperscript{X}]{\includegraphics[scale=0.178]{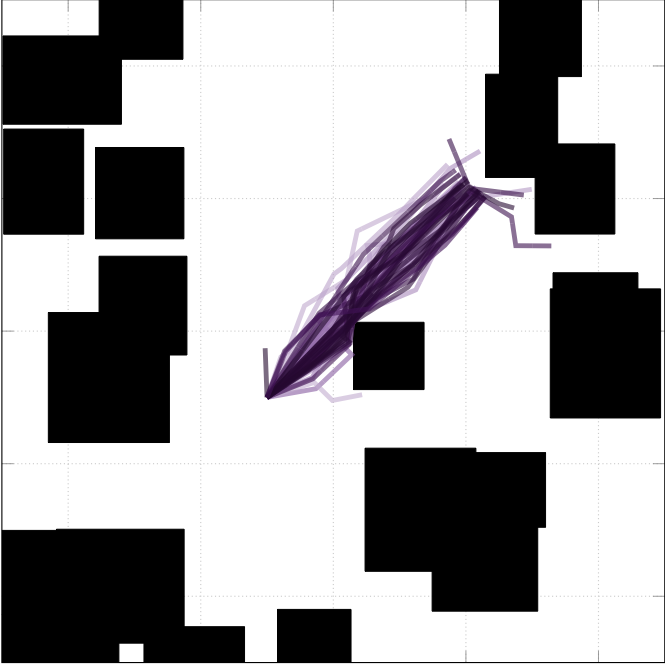}}
  \setlength{\belowcaptionskip}{-10pt}
  \caption{
  A visualization of the paths travelled on 50 attempts of the incremental planning problem done by EIT*, RRT*, RRT-Connect and RRT\textsuperscript{X} with a planning budget of 0.1s. 
  The hue of each path is determined by the order it was run in the experiment. 
  The length of a planner's global solution is the overall length of the path travelled. 
  EIT* found the smallest median global solution length and had a 100\% success rate. 
  RRT* followed similar paths to EIT* due to its ASAO properties, when it was successful, but only succeeded 6\% of the time. 
  RRT-Connect also maintained a 100\% success rate but had a longer median path length than EIT*. RRT\textsuperscript{X} had a 0\% success rate due to the overhead of optimally rewiring its solution trees.
  }
  \label{paths-plot}
\end{figure*}

The quality of the paths found by predictive approaches depends heavily on the quality of the predictions. 
These can be poor in situations with limited preexisting data or highly complex dynamics. In such challenging environments, robots must pair their predictions with reactive planning.

Reactive approaches do not require assumptions about the position or motion of obstacles. 
These methods instead quickly generate intermediate plans when obstacle changes are detected. 
A successful reactive planner must efficiently generate high-quality intermediate paths while remaining consistent with the previously executed path.

Plan-reuse algorithms maintain consistency to previous plans by keeping the existing search and updating the parts affected by the change. 
This can allow them to provide formal guarantees on the optimality of their intermediate solutions (e.g., D* \cite{stentz1994dstar}, LPA* \cite{koenig2004lifelong}, RRT\textsuperscript{X} \cite{otte2015rrtx}). 
They find better global paths than reactive approaches that focus solely on reaction times (e.g., bug algorithms \cite{lumelsky2005bugs}) but can be slower to react to changes in information.

Plan-reuse methods require the locations of obstacle changes in their search trees to efficiently maintain optimal intermediate paths. 
These locations may sometimes be available in the map representation but otherwise require manual evaluation to find. 
Changed edges can always be found by iterating through each edge in the search graph and confirming that it is still obstacle free. 
This increases the time it takes to update plans and may require a prohibitively large number of collision checks on large graphs.

Recent almost-surely asymptotically optimal (ASAO) planners are able to quickly converge towards optimal solutions to complex problems. 
This provides an alternative for fast, globally optimal replanning without the complexity and computational overhead of plan reuse. 
Fast ASAO planners (e.g., Effort Informed Trees; EIT* \cite{Strub_2022}, Asymptotically Optimal RRT-Connect; AORRTC \cite{wilson2025aorrtc}) can instead be run independently each time information changes to quickly find intermediate paths with optimality guarantees.

This challenges the historical assumption that fast incremental planning requires plan reuse \cite{stentz1994dstar} \cite{veloso2002errt} \cite{stentz2006drrt}. 
This paper demonstrates that independent runs of EIT* outperform all other tested incremental planning algorithms on the simulated problems, including RRT\textsuperscript{X}, a planner specifically designed for efficient plan reuse. 
EIT* finds more and higher-quality solutions than all other planners across all tested planning budgets, successfully replanning as quickly as 50ms in simulated experiments. 
It also evaluates this approach on a real robotic platform using AORRTC for moving-obstacle avoidance on a Franka Research 3 arm (Figure \ref{fig:stacked_experiments}).

The rest of this paper is organized as follows. 
Section \ref{lit-review} presents a discussion of current and classic approaches to incremental planning. 
Section \ref{method} presents further background and the independent incremental planning approach. 
Section \ref{experiments} presents the experimental details and Section \ref{discussion} presents a discussion of the results. 
Finally, Section \ref{conclusion} provides a conclusion and discusses future work.

\section{Literature Review} \label{lit-review}

Most replanning systems include a combination of predictive (Section \ref{lit:pred}) and reactive (Section \ref{lit:react}) approaches.

\subsection{Predictive Approaches} \label{lit:pred}

Predictive approaches leverage \textit{a priori} information about their operational environments to inform their path plans but are susceptible to error in highly complex environments. 

Space-Time RRT* (ST-RRT*) \cite{22-grothe-ICRA} uses ASAO planning to find paths through moving obstacles with optimality guarantees when their trajectories are known. 
ST-RRT* uses an ASAO planner in a space augmented with a time dimension to model obstacle movements. 
This approach yields optimal plans through the moving obstacles but assumes perfect knowledge of their trajectories.

Velocity obstacle approaches \cite{fiorini1998velocity} \cite{berg2011accvol} assume that obstacles move up to a predetermined velocity and plan a trajectory to the goal avoiding any region that may be occupied by an obstacle under this constraint. 
Velocity constraints ensure the feasibility of solution paths but conservative bounds may restrict the quality of the solutions found and can create unsolvable problems.

Finite-horizon Model Predictive Control (MPC) can also be used to avoid obstacles. 
These approaches simultaneously react to obstacle changes and predict future trajectories by solving a finite-horizon optimal control problem defined by obstacle predictions. 
This replans quickly but provides no optimality guarantees on the intermediate paths if the problem horizon is shorter than the path to the goal \cite{mpcbook2022}.

\subsection{Reactive Approaches} \label{lit:react}

Reactive approaches do not require assumptions about the dynamics of their environments but must be able to quickly replan to maintain feasibility through the environment.

Bug algorithms \cite{lumelsky2005bugs} avoid expensive planning by following a deterministic policy. Basic bug algorithms drive a straight-line path to the goal when possible and follow the perimeter of any obstacles they encounter. 
Bug algorithms react quickly to new obstacles but provide no optimality guarantees and often execute highly suboptimal paths. 

Following suboptimal intermediate paths may result in the robot unnecessarily changing the homotopy class of its path which can lead to a much longer global solution. 
Many existing approaches avoid this problem by providing guarantees of near-optimal intermediate paths.

Zelinsky \cite{zelinsky1992} presents an algorithm that replans from scratch with A* once environment changes are detected. 
This approach guarantees resolution-optimal intermediate paths but suffers from slow intermediate path formulation.

D* \cite{stentz1994dstar, stentz1995focussed, koenig2002improved, koenig2005fast} extends the A* algorithm to replan when information changes. 
It improves the replanning speed by propagating local graph updates through the previous graph once changes are detected. 
This is proven to find the resolution optimal solution given the information available at the time of planning but assumes it is given all graph changes and does not account for the computational cost of detecting them.
\squeezeWords

Lifelong Planning A* (LPA*) \cite{koenig2004lifelong} also directly extends A* to incremental planning. 
It further improves replanning speed by only propagating updates through edges that are affected by graph changes. 
LPA* also assumes knowledge of which edges in the graph have changed and does not consider the computational cost of detecting changes.

Lifelong Generalized Lazy Search (L-GLS) \cite{lim2024lgls} extends LPA* to take advantage of the efficiency of lazy search algorithms. 
This finds the resolution optimal solution faster than other graph-based incremental replanners by deferring expensive edge checks until necessary but does not avoid the computational costs of detecting changes.

Execution extended RRT (ERRT) \cite{veloso2002errt} extends the Rapidly-exploring Random Trees (RRT) algorithm \cite{lavalle1998rrt} to incremental planning. 
ERRT samples states from previous planning efforts to bias its search toward previous paths. 
This can reduce the replanning time if obstacle locations are similar between planning efforts but provides no guarantees about the quality of its intermediate solutions.

Dynamic RRT (DRRT) \cite{stentz2006drrt} also extends RRTs to incremental planning. 
DRRT reuses feasible branches of previous planning efforts to speed up intermediate path planning but it must validate all of its previous edges once obstacle changes are detected and makes no guarantees about the quality of its intermediate solutions.

RRT\textsuperscript{X} \cite{otte2015rrtx} extends RRT* \cite{karaman2011sampling} to apply ASAO sampling-based planning methods to incremental planning. 
RRT\textsuperscript{X} efficiently repairs its path as information changes and plans from the goal state to avoid unnecessary rerooting of the tree once the planner's location is updated. 
RRT\textsuperscript{X} does not consider the computational costs of detecting which edges in its solution tree are invalidated by obstacle changes.

Multiobjective Dynamic RRT* (MOD-RRT*) \cite{qi2022modrrt} uses a modified RRT* to find an initial path to the goal and then uses a rewiring method to quickly repair paths that collide with obstacles near the robot. 
This rewiring does not explicitly maintain path optimality and therefore provides no guarantees around intermediate path quality. 
MOD-RRT* avoids expensive rewiring cascades by only rewiring locally and not propagating costs through the tree but this reduces the quality of its global solutions.

\subsection{Independent Replanning}

Any single-query planner can be used for incremental planning by solving a new independent planning problem each time new obstacles are detected. 
This does not require specific knowledge of obstacle changes and therefore avoids the computational overhead of many replanning methods. When these problems are solved optimally, the resulting global path will avoid unnecessary changes of direction (i.e., homotopy class switches) but each independent problem must be solved quickly enough to react to obstacle changes.

Recent ASAO planners, such as EIT* \cite{Strub_2022} and AORRTC \cite{wilson2025aorrtc}, quickly find high-quality solutions and  their intermediate paths may be combined into high-quality global solutions without the extra computational cost of replanning.

\section{Methodology} \label{method}

Incremental planning is an extension of optimal planning with imperfect information about the environment.

\subsection{Optimal Planning}

The optimal planning problem is defined similarly to \cite{karaman2011sampling}. Let \(X \subseteq \mathbb{R}^n\) be the state space of the planning problem, 
\(X_{\mathrm{obs}} \subset X\) be the set of states in collision with obstacles, 
and $X_{\mathrm{free}}$ = $\mathrm{closure}(X \setminus X_{\mathrm{obs}})$ be the resulting set of permissible states. 
Let \(x_{\mathrm{start}} \in X_{\mathrm{free}}\) be the initial state and 
\(X_{\mathrm{goal}} \subseteq X_{\mathrm{free}}\) be the set of desired final states. 
Let \(\sigma : [0,1] \to X\) be a sequence of states (i.e., a path) and \(\Sigma\) be the set of all paths.
\squeezeWords

The optimal solution is the path, \(\sigma^*\in\Sigma\), that minimizes a chosen cost function, 
\(c : \Sigma \to \mathbb{R}_{\ge 0}\), while connecting the start, \(x_{\mathrm{start}}\),
to any goal, \(x_{\mathrm{goal}} \in X_{\mathrm{goal}}\), through free space,
\[
\sigma^* 
= \argmin_{\sigma \in \Sigma} 
  \Bigl\{ c(\sigma) \: \Bigm| \: 
    \sigma(0) = x_{\mathrm{start}},\,
    \sigma(1) \in X_{\mathrm{goal}},\]
$$\qquad\quad \forall t \in [0,1],\,
\sigma(t) \in X_{\mathrm{free}}\Bigr\},$$
where $\mathbb{R}_{\ge 0}$ denotes the set of all real numbers greater than or equal to zero.

\subsection{Incremental Planning}

\begin{figure*}[t]
  \centering
  \subfigure[$i=1$]{\includegraphics[scale=0.2]{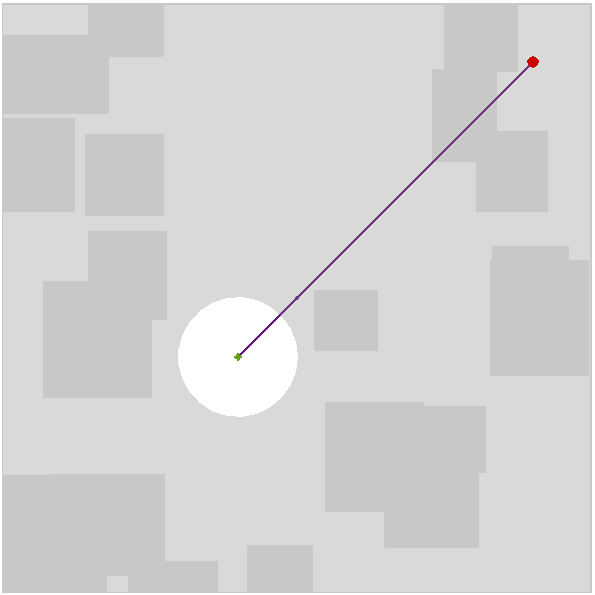}}
  \subfigure[$i=6$]{\includegraphics[scale=0.2]{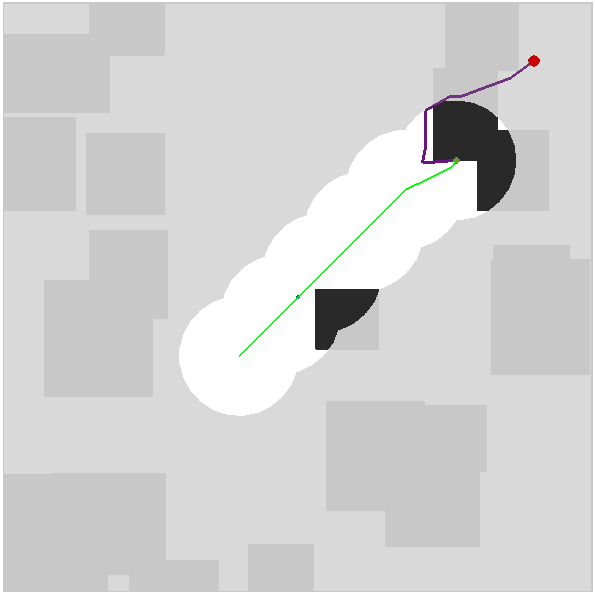}}
  \subfigure[$i=8$]{\includegraphics[scale=0.2]{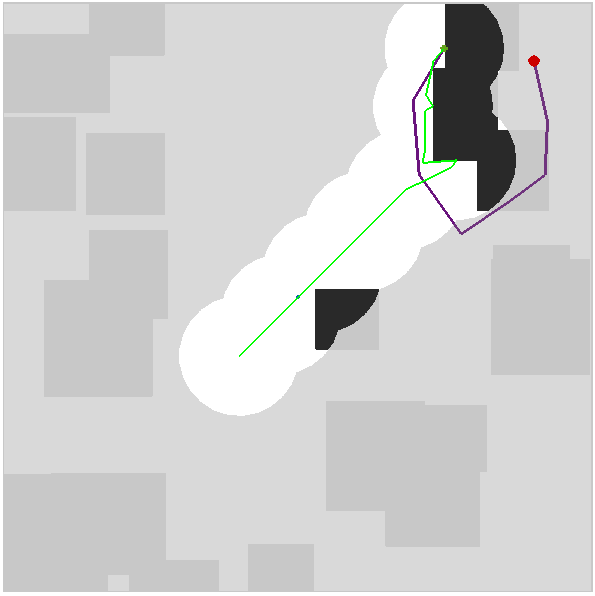}}
  \subfigure[$i=11$]{\includegraphics[scale=0.2]{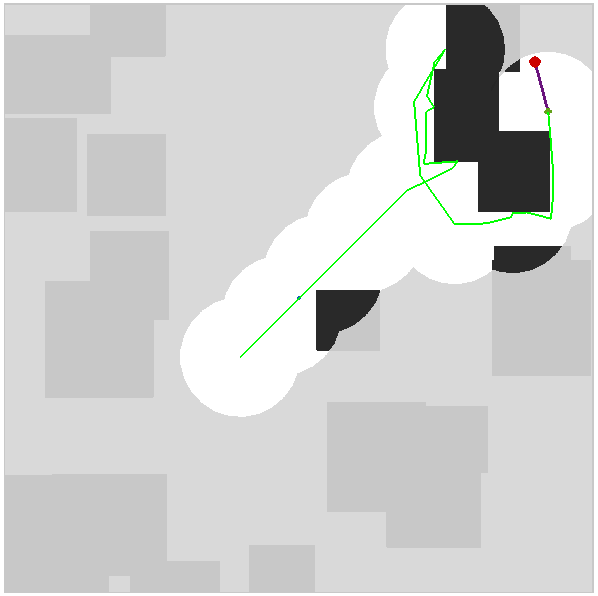}}
  \setlength{\belowcaptionskip}{-10pt}
  \caption{
  A visualization of the intermediate paths found by EIT* with a sensor range of 0.1 and a planning budget of 0.1s. 
  The gray area is unsensed and assumed to be free space by the planner. 
  The dark green dot represents the starting point of the query, and the red dot represents the goal. 
  The green path represents the path travelled so far. 
  The number of circles gives a representation of the number of queries of the planner.
  }
  \label{incremental-plot}
\end{figure*}

The incremental planning problem is an extension of the optimal planning problem where the locations of obstacles in the environment are not perfectly known and may change as the robot moves. 
Exact knowledge of obstacles is assumed to be limited to those detected in a specified distance of the robot, $r_\text{s}$, within a retrospective time horizon, $\tau$.

Let $B(r_\text{s},x)\subset X$ denote a sensing region (e.g., a ball) with maximum sensing range, $r_\text{s} \in \mathbb{R_\mathrm{\ge0}}$, and a center point, $x\in X$. 
The robot has exact knowledge of obstacles that are within the sensor range of its current position, i.e., $X_\mathrm{sensed}(t) = X_\mathrm{obs}(t) \cap B(r_\mathrm{s},x)$, where $X_\mathrm{obs}(t)$ is the obstacles at time $t$. Let $X_{\mathrm{sensed},i}$ be the set of all obstacles sensed by the robot and considered for planning at the $i^\text{th}$ iteration, $X_{\mathrm{sensed},i} = \bigcup_{t=t_i}^{t_j} X_\mathrm{sensed}(t)$, where $t_{i} = t_j-\tau$ is the minimum time from which the planner still uses its previous obstacle measurements. 
The robot may also include a prior over obstacle motions in its set of obstacles.

The robot plans in the free space defined by the set of obstacles sensed within the retrospective time horizon, $X_{\mathrm{free},i} = X \setminus X_{\mathrm{sensed},i}$, and follows the resulting solution path, $\sigma_i$, until it reaches a state at which it determines it must replan from, $x_{i+1}$. This process repeats until the robot reaches the goal, $X_\mathrm{goal}$.

Let $s_i \in [0,1]$ denote the parameter at which the robot replans from when following its $i^\text{th}$ intermediate solution path, i.e., $x_{i+1} = \sigma_i(s_i)$. Let \( \sigma_{i, s_i} : [0, s_i] \to X \) denote the partial path travelled during the \(i^\text{th}\) planning iteration. 
The global solution path is defined as \( \pi = \sigma_{1,s_1} \oplus \sigma_{2, s_2} \oplus \cdots \oplus \sigma_{N, s_N} \), where \( \oplus \) denotes path concatenation and \( N\in\mathbb{Z_\mathrm{\ge0}} \) is the number of planning cycles until the goal is reached. 
An incremental planner seeks to minimize the cost of its global solution, $c(\pi)$.

\subsection{Independent Incremental Planning Approach}

The independent incremental planning approach uses independent calls to ASAO planners to solve the incremental planning problem. 
The robot solves an independent optimal planning problem in the free space defined by the sensed obstacles at each planning iteration, $X_{\mathrm{free}, i} = X\setminus X_{\mathrm{sensed}, i}$, to find a path, $\sigma_i$. 
The robot follows the path until it arrives at a state at the edge of the sensed area or otherwise determines it needs to replan, $x_{i+1}$. 
The robot updates its obstacles using the sensor and then replans from the new position. The process of planning, following the solution path and sensing repeats until the robot reaches the goal. 

Let $\sigma^{*}_i$ be the shortest path in the known free space from the current state, $x_i$, to the goal. 
Let $\sigma^{*}_{i, s_i}$ denote the subpath of $\sigma^{*}_i$ that ends at the next sensing boundary. 
Following the optimal path with respect to the given information at each iteration, $\sigma^{*}_{i, s_i}$, minimizes the worst-case total path cost over all possible environments, given the current information.

The quality of a global solution to the incremental planning problem depends on the quality of its intermediate paths. 
If the planner finds sufficiently optimal intermediate solutions then the executed path will avoid oscillating between different homotopy classes (i.e., be consistent) without explicitly considering path consistency and result in a near optimal global solution. If the planner does not provide optimality guarantees then it may yield suboptimal intermediate paths that are inefficient, backtrack, and/or unnecessarily switch homotopy classes and generally reduce the quality of the global solution.

\begin{algorithm}[t]\small
\caption{Independent Incremental Planning}
\label{alg:inc}
\SetAlgoLined
\KwIn{$X$, $x_0$, $x_{\mathrm{goal}}$, $r_{\mathrm{s}}$, \texttt{planner}}

$\Xsense{0}\!\leftarrow\!\Xobs 0 \cap \Ball{r_{\mathrm{s}}}{x_0}$\;
$\Xfree{0}\!\leftarrow\!X\setminus\Xsense{0}$\;
$\pi\!\leftarrow\!\emptyset$; $x_i\!\leftarrow\!x_0$\;

\Repeat{$x_i = X_{\mathrm{goal}}$}{
  $\sigma_i \!\leftarrow\! \texttt{planner}\big(x_i,\,x_{\mathrm{goal}},\Xfree{i}\big)$\;
  $\pi \!\leftarrow\! \pi \oplus \sigma_{i,s_i}$\;
  $x_{j} \!\leftarrow\! \sigma_i(s_i)$\;
  $c(\pi) \!\leftarrow\! c(\pi)+c(\sigma_{i,s_i})$\;

  $X_{\mathrm{sensed},i} = \bigcup_{t=t_i}^{t_j} X_\mathrm{sensed}(t)$\;
  $\Xfree{i}\!\leftarrow\!X\setminus\Xsense{i}$\; $x_i\!\leftarrow\!x_j$\;
}
\Return{$\pi$}
\end{algorithm}

\section{Experiments} \label{experiments}

\begin{figure*}[t]
\centering

\begin{minipage}{0.31\textwidth}
  \centering
  \subfigure[Random Rectangles]{\includegraphics[width=0.7\linewidth]{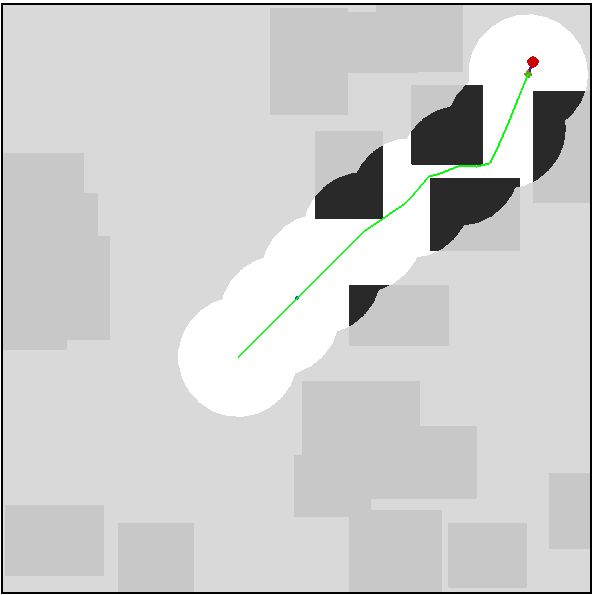}}\\[0.6ex]
    \subfigure[Random Rectangles Success Rates at 100ms]{\includegraphics[width=\linewidth]{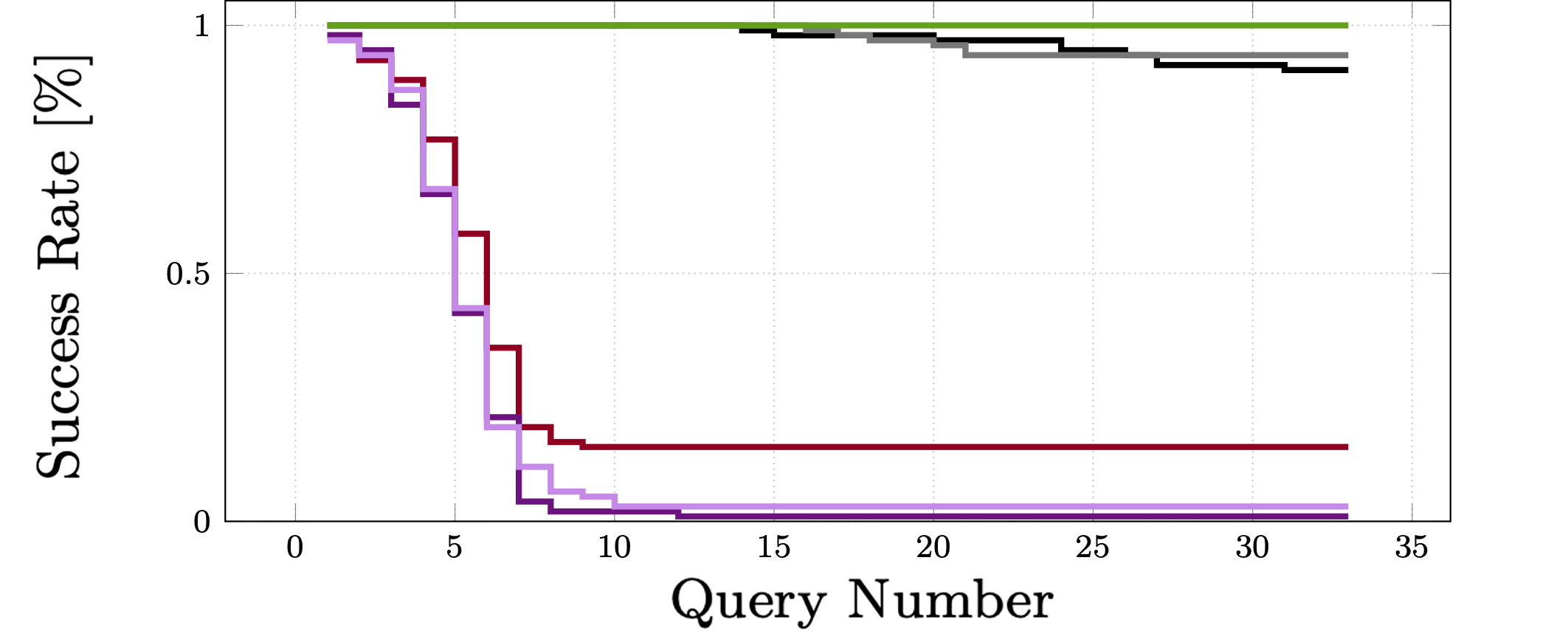}}\\[0.6ex]
  \subfigure[Random Rectangles Success Rates at 50ms]{\includegraphics[width=\linewidth]{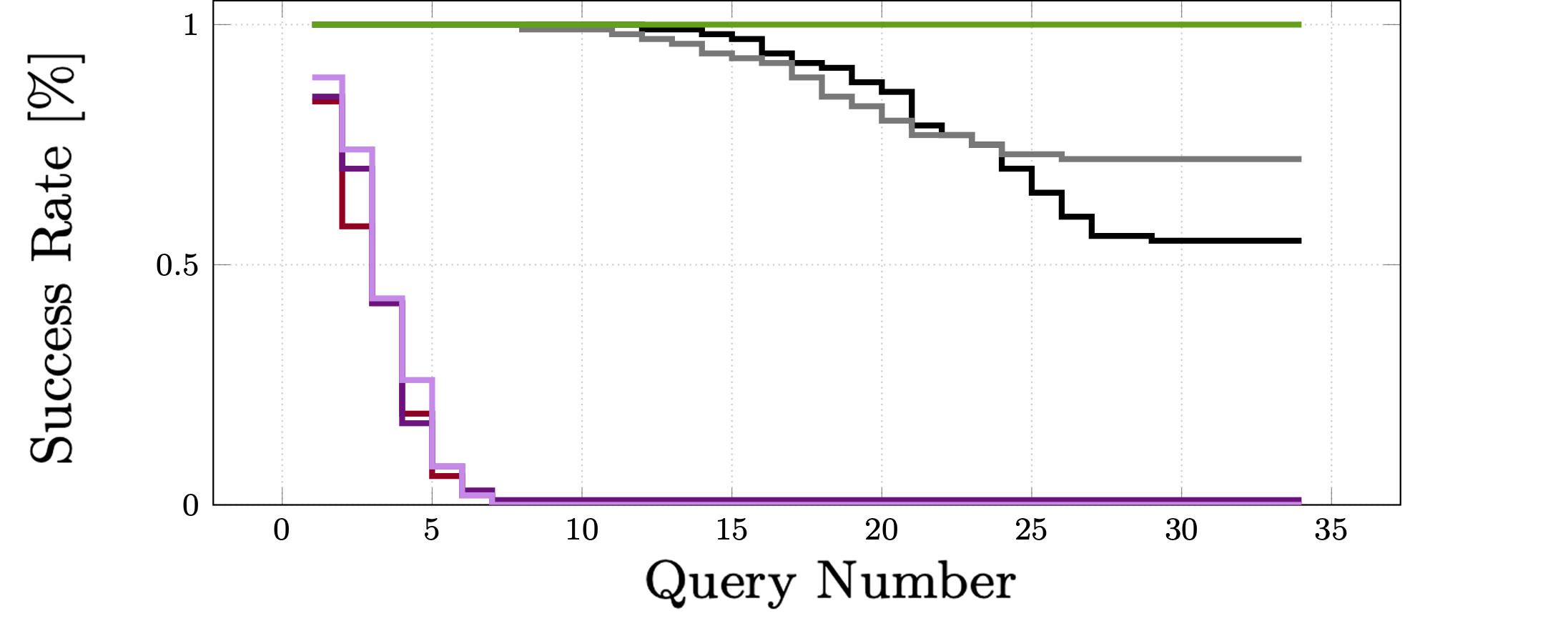}}\\[0.6ex]

\end{minipage}\hfill
\begin{minipage}{0.31\textwidth}
  \centering
  \subfigure[Wall Gap]{\includegraphics[width=0.7\linewidth]{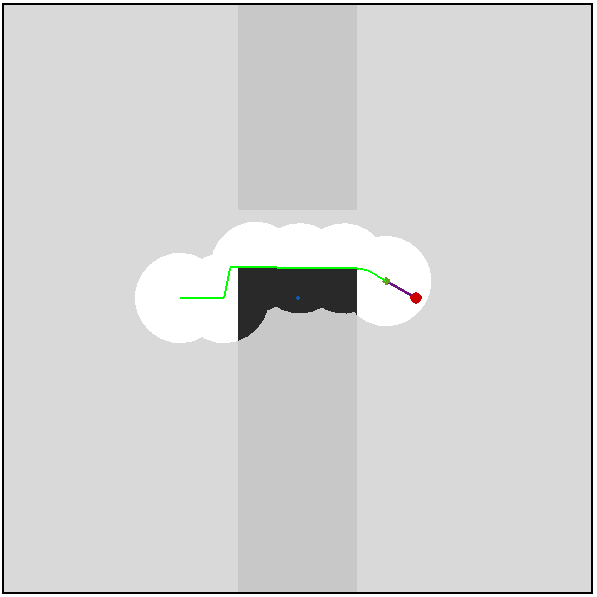}}\\[0.6ex]
  \subfigure[Wall Gap Success Rates at 100ms]{\includegraphics[width=\linewidth]{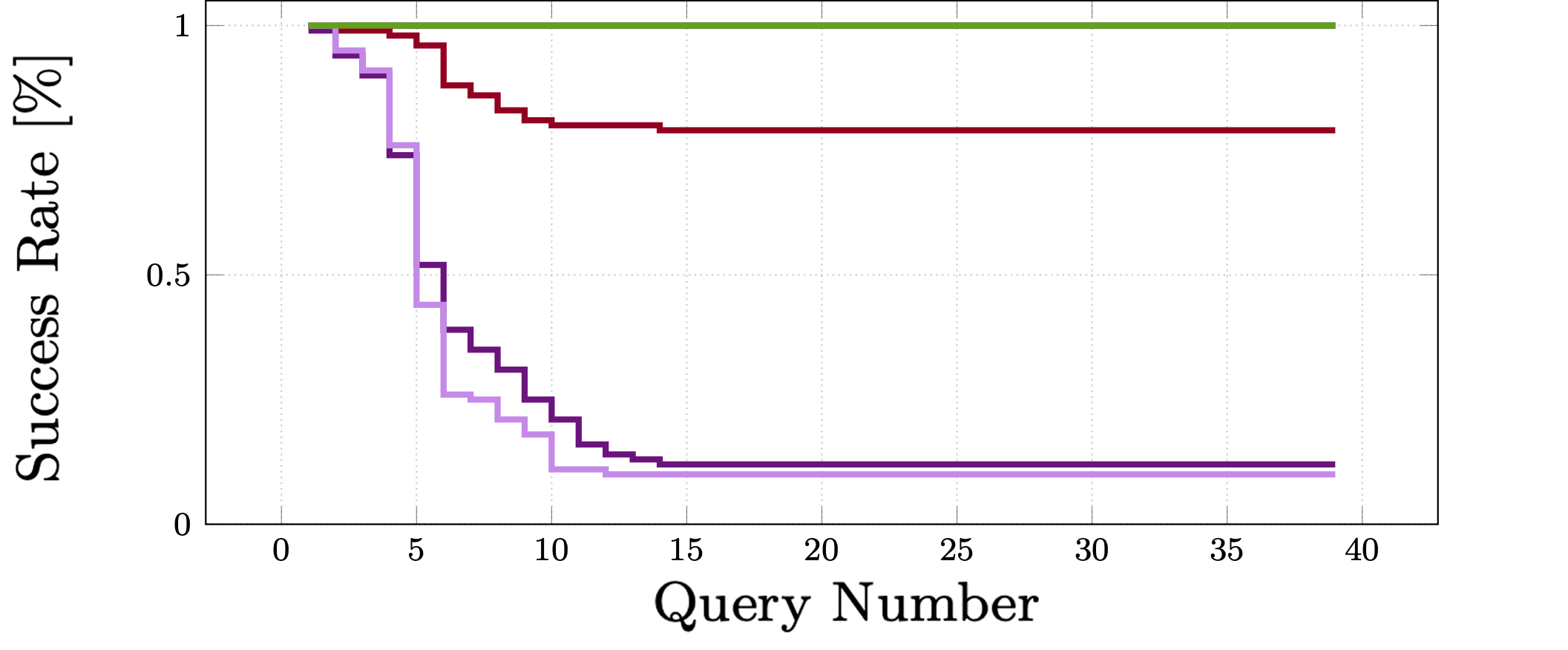}}\\[0.6ex]
  \subfigure[Wall Gap Success Rates at 50ms]{\includegraphics[width=\linewidth]{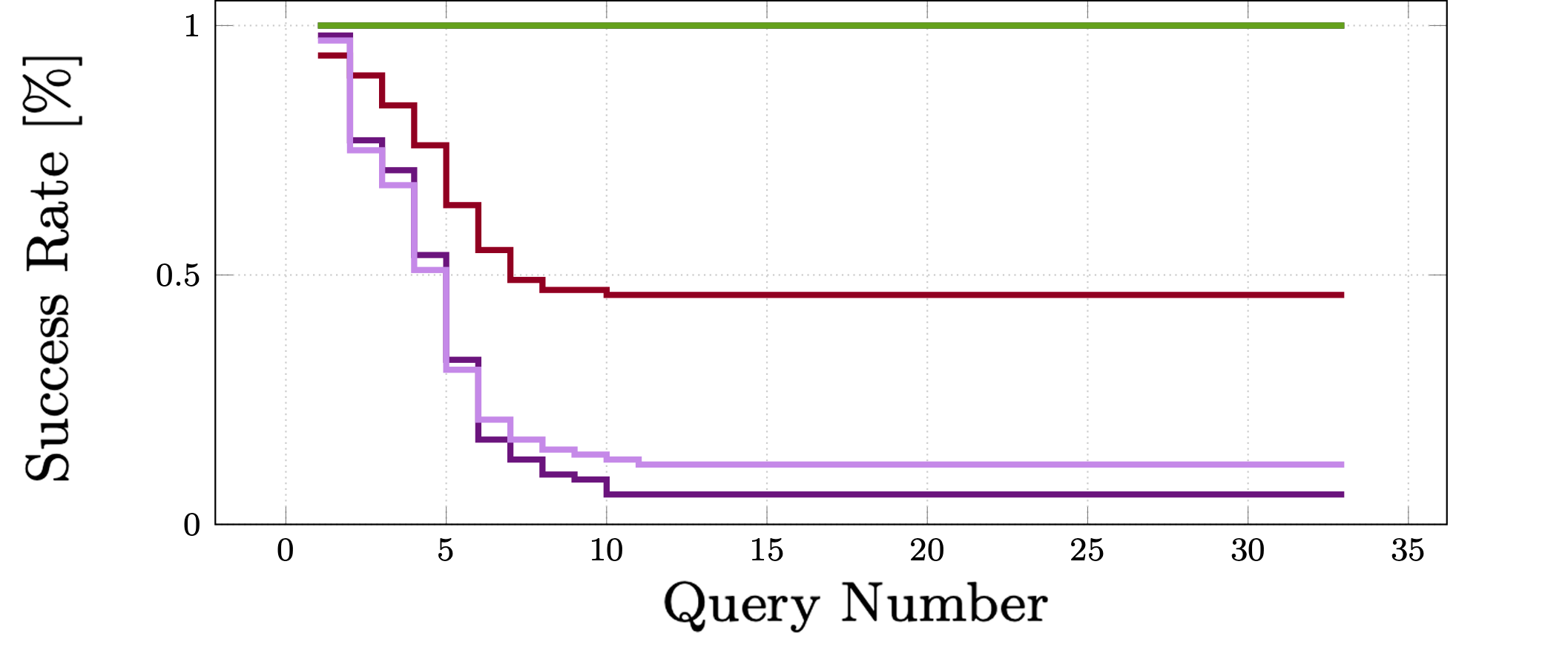}}\\[0.6ex]

\end{minipage}\hfill
\begin{minipage}{0.31\textwidth}
  \centering
  \subfigure[Double Enclosure]{\includegraphics[width=0.7\linewidth]{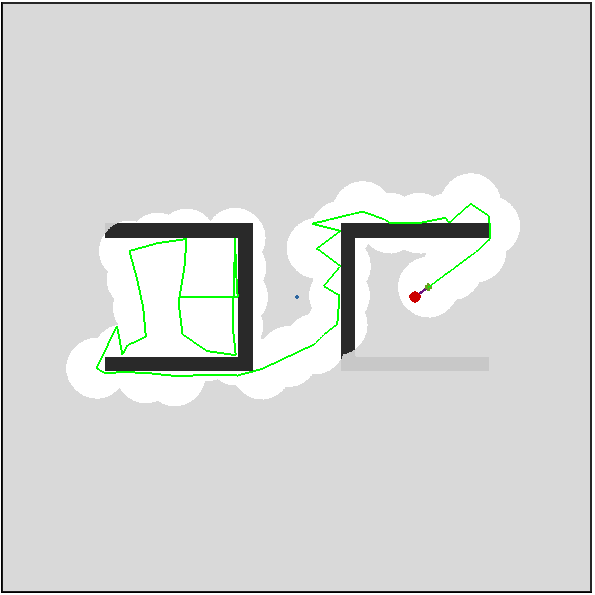}}
  \subfigure[Double Enclosure Success Rates at 100ms]{\includegraphics[width=\linewidth]{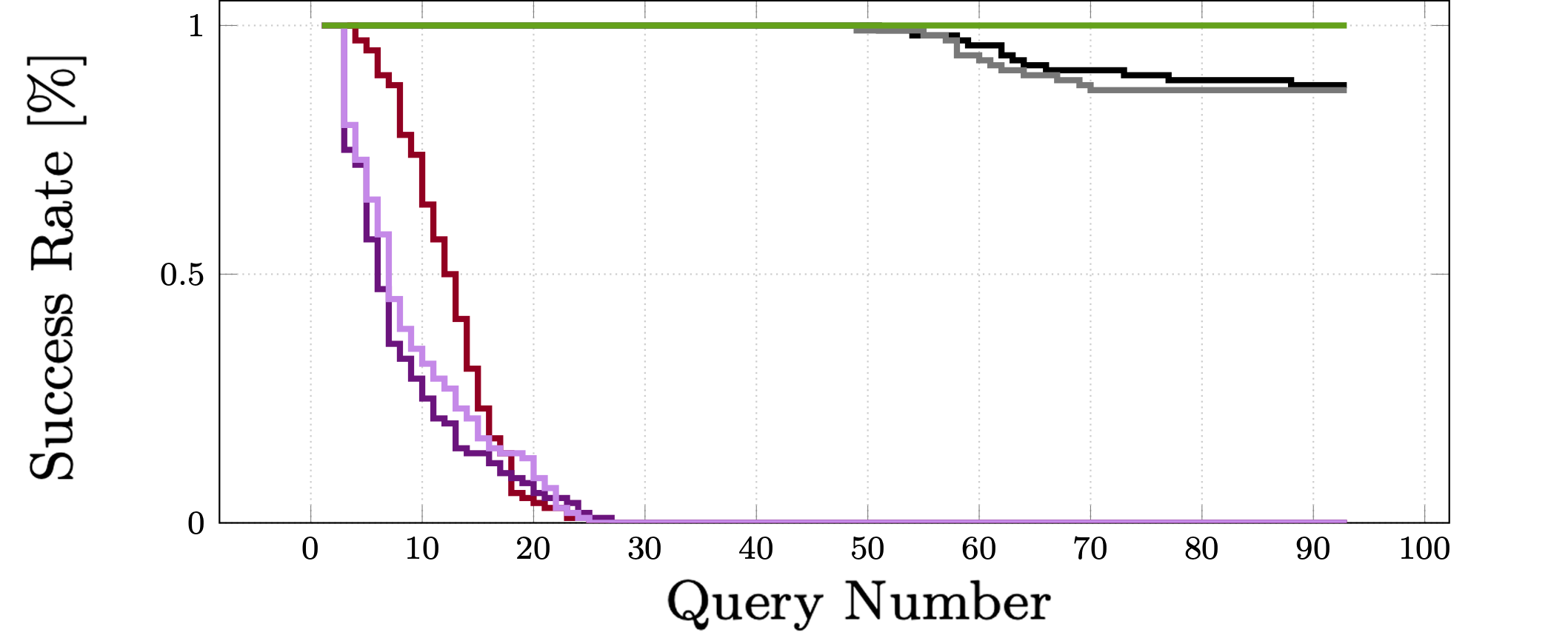}}
  \subfigure[Double Enclosure Success Rates at 50ms]{\includegraphics[width=\linewidth]{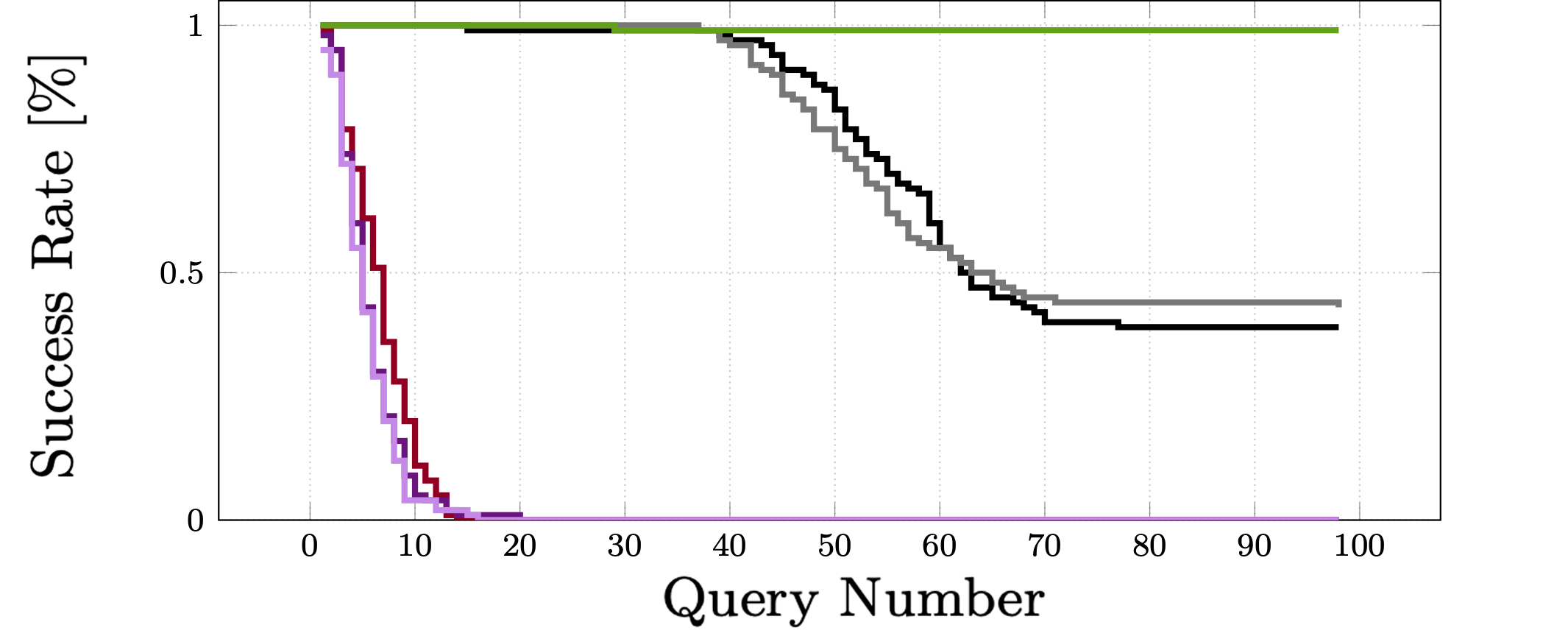}}
\end{minipage}

\includegraphics[scale=0.25]{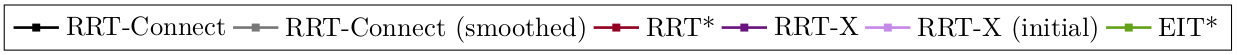}

\setlength{\belowcaptionskip}{-10pt}

\caption{The success rates of all planners across 100 runs on three of the simulated worlds. A visualization of the world as well as a solution found by EIT* is shown in (a), (d) and (g). Each of the planners was tested with a planning budget of $0.1s$ and $0.05s$. The success rate at the $n^\text{th}$ query is defined by the percentage of planners that are able to find a solution to the $n^\text{th}$ intermediate problem within the planning budget. Success rates are carried forward, i.e., if a planner fails on a query it is also considered to fail on all subsequent queries. The independent ASAO approach with EIT* maintained the highest success rate across every world and every planning budget while finding the shortest global solution (Table \ref{tab:results}).}
\label{success-rates}
\end{figure*}

The independent incremental planning approach was tested on both probabilistically complete (RRT-Connect \cite{kuffner2000rrtconnect} with and without smoothing) and ASAO planners (RRT* and EIT*) and was compared against a state-of-the-art planner designed for incremental replanning problems (RRT\textsuperscript{X}). Since the planning time of RRT\textsuperscript{X} is slow when it has to optimally rewire dense solution trees, results are presented for both the full version of RRT\textsuperscript{X} and a version that stops after it finds an initial solution to provide best-case results for RRT\textsuperscript{X} with respect to initial solution time.

Each of the planners were tested on three types of simulated worlds using publicly available Open Motion Planning Library (OMPL) \cite{sucan2012ompl} implementations with the partial RRT\textsuperscript{X} implementation publicly available in OMPL extended to handle replanning. EIT* used a batch size of 100, radius factor of 1.001, repair factor of 1.2 and used the $k$-nearest implementation with pruning. RRT-based planners used a maximum edge length of 0.3 and single-tree RRT-based planners used a goal bias of 0.05. RRT\textsuperscript{X} was run without informed sampling and an epsilon of zero. RRT* was run with a rewire factor of 1.001. 

The incremental planning problem was simulated as an extension of the Planner Development Tools (PDT) \cite{gammell_empp22}. The simulation requires three parameters to describe the environment: a planning budget, $t$, a maximum sensor range, $r_\text{s}$, and a global problem. The environment the planner interacts with for each intermediate problem is referred to as the \emph{incremental} environment. The simulation assumes that all sensed obstacles are stationary, i.e., $\tau=\infty$. 

The incremental environment starts out empty with the same start, $x_0$, and goal, $X_\text{goal}$, as the global environment. Only the obstacles within the sensor range from the start are added, and the planner attempts to find a path from the start to the goal that avoids these obstacles before the planning budget runs out. If the planner succeeds in finding a solution, the robot follows this path until it reaches a state at the end of the sensor radius, $x_{1}$. Once the robot has moved, a circle centered at its current position with a radius equal to $r_\text{s}$, i.e., $\Ball{r_{\mathrm{s}}}{x_{1}}$, is added to the sensed area. The next incremental problem is defined as a new optimal planning problem using $x_{1}$ as its starting point. The obstacles in this problem are the intersection of all obstacles in the global environment and the (now larger) sensed area. This process is repeated until the planner reaches the goal (Figure \ref{incremental-plot}). 
This sequential planning problem under the free space assumption occurs when planning with limited sensing range and also for dynamic obstacles without any predictions.

The performance of RRT\textsuperscript{X} depends on the computational cost of detecting edges that are invalidated by newly discovered obstacles. This can be computationally inexpensive in map representations that provide this information directly or expensive when each existing edge must be checked each time new obstacle information is received. In order to provide best-case results for RRT\textsuperscript{X}, the computational cost of detecting invalidated edges for RRT\textsuperscript{X} was ignored in the experiments.

\subsection{Simulated Experiments}

The simulated worlds are two-dimensional path length minimization problems in a square environment, $X=[-1, 1]^2$. A sensor range of 0.1 was used on each Random Rectangles world, 0.075 on the Wall Gap world and 0.05 on the Double Enclosure world.

\subsubsection{Random Rectangles}

Each random world consists of 20 randomly placed rectangles with side lengths uniformly distributed between 0.1 and 0.2. The starting state is $(-0.1, -0.1)$, and the goal state is $(0.4, 0.4)$. Each world is checked for validity (i.e., a feasible path exists) before it is used. Figure \ref{success-rates}a illustrates a representative world for which individual results are presented. The random worlds evaluate the capability of each planner to quickly navigate around many obstacles to reach the goal.

\subsubsection{Wall Gap}

The wall gap world (Figure \ref{success-rates}d) has two rectangles as obstacles centered about the y-axis. There is a narrow rectangular gap between the two rectangles that the planner must navigate through. The wall gap tests the ability for planners to find and navigate through a narrow gap at the center of the world.

\subsubsection{Double Enclosure}

The double enclosure is two rectangular enclosures, each with one open end, containing the start and goal (Figure \ref{success-rates}g). 
The planner initially sees an unobstructed path from start to goal but will discover it is blocked after travelling towards it. 
The planner must then retrace back towards the start to leave the first enclosure and then circumnavigate the second enclosure to reach the goal. 
The double enclosure tests the planner's ability to adapt when initial plans are proven invalid by new information.

\begin{figure}
    \centering
    \includegraphics[width=\linewidth]{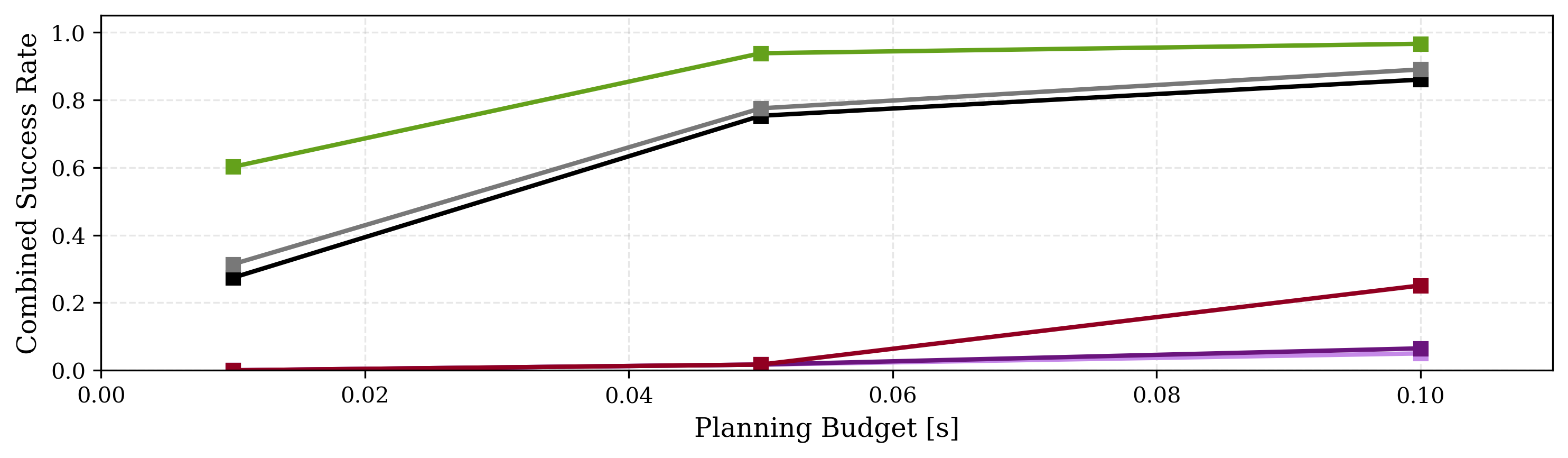}
    \includegraphics[width=\linewidth]{new-legend.png}
    \setlength{\belowcaptionskip}{-14pt}
    \caption{The combined success rate of each planner on 100 different random worlds for planning budgets of 0.01s, 0.05s and 0.1s. EIT* had the highest total success rate at each planning budget and RRT\textsuperscript{X} was not able to consistently solve the problems in the available time.}
\label{fig:success_vs_budget}
\end{figure}

Three sets of experiments were run using the simulated worlds. Every planner was run 100 times on the Wall Gap problem, Double Enclosure problem and 40 different Random Rectangles problems with a planning budget of 0.1s to investigate the performance of each planner on a diverse set of problems. Every planner was run 100 more times on the Wall Gap problem, Double Enclosure problem and a representative Random Rectangles problem with a planning budget of 0.05s to investigate their performance on representative problems with a smaller budget. Each planner was also run 10 times with planning budgets of 0.01s, 0.05s and 0.1s on 100 different Random Rectangles problems for each planning budget to investigate the performance of each planner at different planning budgets (Figure \ref{fig:success_vs_budget}).

\subsection{Results}

The per-query success rate, global solution time, global path length, and number of queries were measured for each simulated world. The combined success rate, median problem success and median rank of each planner's median global path length, global solution time, and number of queries were also collected across the 40 random worlds. The combined success rate across all runs over the 100 random worlds was collected for each planning budget. A summary of the results from a planning budget of 0.1 seconds is presented in Table \ref{tab:results}. Only the success rates are shown for the 50ms experiments (Figure \ref{success-rates}) since EIT* was the only planner that solved more than 50\% of all trials on all of the worlds. A plot of the total success rate over 100 random worlds at each planning budget is shown in Figure \ref{fig:success_vs_budget}.

\subsubsection{Success Rates}

The success rate for each query is shown in Figure \ref{success-rates}. The success rate at a given query is a measure of how often the planner is able to find a solution within the given planning budget at that query. The global success rate (i.e., the percentage of runs that the planner reached the goal) can be extracted from the success rate at the final query. EIT* had the highest global success rate on all simulated worlds at all evaluated planning budgets. None of the evaluated planners had 100\% success on all problems because some randomly generated worlds require solutions to pass through very narrow gaps to reach the goal. Increasing the planning budget further on these difficult problems would increase their success rate.  

\subsubsection{Global Path Length}

The median global path length is shown in Table \ref{tab:results}. The global path length measures how far each planner travelled from the start to the goal. The global path length is the length of the portions of the intermediate paths followed to travel between the start and goal. EIT* had the shortest median global path length on all problems due to the high-quality intermediate paths it finds as a result of its ASAO properties.

\subsubsection{Global Solution Time \& Queries}

The median global solution time and median number of queries are shown in Table \ref{tab:results}. The global solution time is the total time spent finding a solution to the incremental planning problem. The number of queries is the number of intermediate queries a planner took to reach the goal.

Nonoptimal planners, such as RRT-Connect, do not spend the whole planning budget and instead return as soon as they find an initial solution. Optimal planners, such as RRT* and RRT\textsuperscript{X}, first find an initial solution and then spend the rest of the planning budget searching for shorter paths. Other optimal planners, such as EIT*, have the capability to exit early if they detect that they have found the optimal solution. The number of queries is presented alongside the global solution time to give a fair comparison between planners.

EIT* had the smallest median number of queries on all problems. RRT-Connect had a smaller median global solution time than EIT* on the Wall Gap and Double Enclosure problems since it does not use the whole planning budget. EIT* maintained the smallest median global planning budget on the selected Random Rectangles problem despite spending the whole planning budget optimizing its paths due to the small number of queries it needs to solve the problem.

\begin{table*}[t]
\caption{
A summary of all planning results on the Random Rectangles, Wall Gap and Double Enclosure simulated worlds. 
Each result shows the percentage of successful runs, the median solution path length, the median solution time, and the median number of queries taken to reach the goal. 
Clopper-Pearson 95\% confidence intervals are shown in brackets for single-environment success rates and nonparametric 95\% confidence intervals for the median of all other metrics.
The best result on each environment is shown in bold. 
Unsuccessful trials were considered to have infinite path length, solution time, and queries. 
The combined success rate, median success rate and the median rank of each planner's median path length, median global solution time and median number of queries are shown for the 40 different Random Rectangles worlds.
}
\centering
\setlength{\tabcolsep}{3pt}
\renewcommand{\arraystretch}{1.15}
\footnotesize
 
\NewDocumentCommand{\CIs}{m m}{{\scriptsize[#1, #2]}}
\begin{tabular}{p{1.4cm}ccccccc}
\toprule
& & \multicolumn{6}{c}{\textbf{Planner}} \\
\cmidrule(lr){3-8}
\textbf{Environment} & \textbf{Metric}
& RRT-Connect
& RRT-Connect (smoothed)
& RRT*
& RRT\textsuperscript{X}
& RRT\textsuperscript{X} (initial)
& EIT* \\
\midrule
\multirow{4}{*}{\thead[l]{\textbf{Rand. Rect.} \\ \textbf{(Example)}}}
& Succ. Rate   & 91\% \CIs{83.6}{95.8} & 94\% \CIs{87.4}{97.8} & 9\% \CIs{8.6}{23.5} & 1\% \CIs{0.0}{5.4} & 3\% \CIs{0.6}{8.5} & \textbf{100\%} \CIs{96.4}{100.0} \\
& Path Len.    & 4.78 \CIs{4.15}{5.34} & 3.14 \CIs{2.93}{3.34} & $\infty$ & $\infty$ & $\infty$ & \textbf{0.718} \CIs{0.717}{0.719} \\
& Sol. Time  & 0.366s \CIs{0.291}{0.388} & 0.540s \CIs{0.451}{0.638} & $\infty$ & $\infty$ & $\infty$ & \textbf{0.313s} \CIs{0.313}{0.313} \\
& Queries        & 23 \CIs{21}{25} & 18 \CIs{17}{19} & $\infty$ & $\infty$ & $\infty$ & \textbf{7} \CIs{7}{7} \\
\midrule
\multirow{4}{*}{\textbf{Wall Gap}}
& Succ. Rate   & \textbf{100\%} \CIs{96.4}{100} & \textbf{100\%} \CIs{96.4}{100} & 79\% \CIs{69.7}{86.5} & 12\% \CIs{6.4}{20.0} & 10\% \CIs{4.9}{17.6} & \textbf{100\%} \CIs{96.4}{100} \\
& Path Len.    & 1.95 \CIs{1.65}{2.15} & 1.72 \CIs{1.37}{1.84} & 0.543 \CIs{0.522}{0.615} & $\infty$ & $\infty$ & \textbf{0.389} \CIs{0.388}{0.390} \\
& Sol. Time  & \textbf{0.070s} \CIs{0.054}{0.074} & 0.073s \CIs{0.062}{0.087} & 0.705s \CIs{0.608}{0.706} & $\infty$ & $\infty$ & 0.399s \CIs{0.399}{0.399} \\
& Queries        & 15 \CIs{15}{18} & 13 \CIs{13}{15} & 7 \CIs{6}{7} & $\infty$ & $\infty$ & \textbf{5} \CIs{5}{5} \\
\midrule
\multirow{4}{*}{\thead[l]{\textbf{Double} \\ \textbf{Enclosure}}}
& Succ. Rate   & 88\% \CIs{80.0}{93.6} & 87\% \CIs{78.8}{92.9} & 0\% \CIs{0.0}{3.6} & 0\% \CIs{0.0}{3.6} & 0\% \CIs{0.0}{3.6} & \textbf{100\%} \CIs{96.4}{100} \\
& Path Len.    & 6.32 \CIs{5.90}{6.90} & 6.05 \CIs{5.69}{6.52} & $\infty$ & $\infty$ & $\infty$ & \textbf{2.01} \CIs{1.99}{2.02} \\
& Sol. Time  & \textbf{1.14s} \CIs{1.09}{1.21} & 1.88s \CIs{1.72}{2.10} & $\infty$ & $\infty$ & $\infty$ & 2.50s \CIs{2.40}{2.50} \\
& Queries        & 70 \CIs{67}{72} & 66 \CIs{63}{68} & $\infty$ & $\infty$ & $\infty$ & \textbf{31} \CIs{31}{31} \\
\midrule
\multirow{3}{*}{\thead[l]{\textbf{Rand. Rect.} \\ \textbf{(Agg.)}}}
& Total Succ.   & 95.2\% & 96.0\% & 26.2\% & 8.13\% & 8.70\% & \textbf{98.9\%} \\
& Med. Succ.  & 99.0\% \CIs{99.0}{100} & \textbf{100\%} \CIs{99.0}{100} & 18.0\% \CIs{11.0}{35.0} & 4.0\% \CIs{1.0}{8.0} & 2.0\% \CIs{1.0}{9.0} & \textbf{100\%} \CIs{100}{100} \\
& Med. Rank       & 3/\textbf{1}/3 & 2/2/2 & 4/4/4 & 4/4/5 & 4/4/6 & \textbf{1}/3/\textbf{1} \\
\bottomrule
\end{tabular}

\label{tab:results}
\end{table*}

\subsection{Real-World Planning}

A VAMP \cite{thomason_motions_microseconds_2024} implementation of AORRTC was used as the incremental planner to navigate a seven DoF Franka Research 3 robot arm around moving obstacles. A collision-affording point tree (CAPT) \cite{Ramsey-RSS-24} was used to efficiently map obstacles presented in different configurations and AORRTC planned a new solution around each of the different obstacles with no prior knowledge of their locations.

AORRTC successfully navigated past each obstacle configuration in real-time due to its ability to quickly find high-quality intermediate paths. The robotic arm experiment demonstrates the ability for fast ASAO planners to navigate successfully in high-dimensional environments with moving obstacles without modelling their motion.

\vspace{-0.2em}
\section{Discussion} \label{discussion}

The simulated experiments show that the incremental planning problem can be solved by independent planning and that when these planners find high-quality solutions quickly this independent incremental planning approach outperforms specifically designed incremental planners.

RRT-Connect was able to quickly find solutions to intermediate problems but had larger median global path lengths than the ASAO planners. The lack of solution quality guarantees meant that RRT-Connect did not find consistent paths, both with and without smoothing. This resulted in paths that unnecessarily left the sensed area or changed homotopy classes and resulted in more planning queries than necessary. The smoothed variation of RRT-Connect outperformed unsmoothed RRT-Connect on path length and number of queries but had a longer median path length and larger number of queries than EIT* across all problems. Both variations of RRT-Connect had lower total success rates than EIT* on all sets of random worlds because their inconsistent intermediate solutions cause them to solve an unnecessarily large number of planning queries to reach the goal.

RRT\textsuperscript{X} failed to find a complete global solution within the incremental planning budget on more than 90\% of its trials across all simulated worlds. This is because RRT\textsuperscript{X} spends computational effort to update its entire search tree each time obstacles change, even if parts of it are not necessary to solve an individual query. RRT\textsuperscript{X} has to rewire potentially very large trees when there are significant obstacle changes that affect many edges in its solution tree, which is computationally expensive. 

The computational cost of this tree update also increases with the planning budget. A large planning budget means that more nodes will be inserted during each query and therefore need to be rewired and collision checked when updating the tree in future queries. This is an inherent trade-off of anytime-resolution plan-reuse algorithms, since the tree becomes more dense with more time between queries. This has been addressed in multiquery settings by rewinding the sample density between queries \cite{Hartmann_2023}. 

The computational cost of optimally rewiring dense solution trees is further demonstrated by the performance of RRT\textsuperscript{X} when it is limited to only finding an initial solution. This truncated version of RRT\textsuperscript{X} was able to find solutions to more intermediate queries than the full RRT\textsuperscript{X} on most simulated experiments but failed to find a complete global solution within the incremental planning budget on more than 90\% of the trials.

RRT* was able to find shorter median global path lengths than RRT-Connect and RRT\textsuperscript{X} but struggled to find initial solutions on small planning budgets. The ASAO properties of RRT* ensure that when it is given sufficient planning time it converges towards the optimal solution. This consistency between intermediate solutions ensures that RRT* only exits the sensed area when necessary but the required planning budget proved to be large. RRT* failed more than half the time on the Double Enclosure problems and all sets of Random Rectangles due to its slow initial solution time.

EIT* is able to find fast and near-optimal intermediate solutions since it is designed to quickly find an initial solution and then converge toward the optimal path. EIT* found shorter median solution paths than all other tested planning algorithms at all planning budgets. EIT* had the highest success rates on all sets of Random Rectangles problems at every planning budget and was the only planner to solve more than 50\% of the problems on all experiments.

The real-world experiments demonstrate the applicability of the independent ASAO replanning strategy to other fast ASAO planning algorithms, dynamic obstacles, and real-world robotics problems. Suitably fast ASAO planners are able to replan each time obstacle changes are detected and achieve real-time dynamic planning, as demonstrated in the Franka arm experiment. 

\pushLines{
The simplicity of implementing the independent ASAO approach is also worth noting. Replanning from scratch simplifies many modeling assumptions in dynamic environments by not relying on a prior or attempting to reuse information between queries. An incremental planner that tries to reuse information from previous planning efforts would incur a performance overhead on long-horizon planning problems like the Franka arm experiment and would require careful tuning or sophisticated methods to control its solution-tree density. Many existing dynamic planners would require prior information about their obstacles like a maximum velocity or probabilities of different obstacle trajectories to navigate in the experiment. The only parameter tuning needed to run the independent ASAO replanning strategy is to use the largest budget that can be afforded and no problem-specific information or prior data is required, although it can be incorporated into the obstacle map if available.
}

\section{Conclusion and Future Work} \label{conclusion}

Traditional reactive planning literature focuses on trying to reuse information from previous queries in order to speed up planning time. This is often done by attempting to repair old plans when new obstacle information becomes available in order to maintain their path validity, consistency, and/or optimality. For this to be computationally tractable, many current algorithms assume that there is an efficient method to identify the locations of obstacle changes and propagate this information throughout their existing plans.

This paper shows that independent calls to ASAO planners can instead outperform these planners on incremental planning problems. We demonstrate that a suitably fast ASAO planner, such as EIT*, outperforms information-reuse algorithms, such as RRT\textsuperscript{X}, even when access to obstacle changes is free. It also shows that EIT* finds lower-cost solutions more quickly and with a higher success rate than independent runs of other ASAO algorithms, such as RRT*, and fast probabilistically complete algorithms like RRT-Connect.

The results of this paper demonstrate the capability of ASAO planners to replan in simulation as quickly as 50ms and the applicability of this approach to a real-world robotic arm problem. This capability is exciting as it shows the ability to find near-optimal global solutions at close to the speed of control-level systems. In the future we would like to investigate using the independent ASAO approach on robots with obstacle predictions to find better solutions in environments with dynamic obstacles and on robots with kinodynamic or other constraints.

\section{Acknowledgements}

Sneha Ramshanker did an early investigation for this work at the University of Oxford. This work was supported by a Charles Allan Thompson Award and the Natural Sciences and Engineering Research Council of Canada (NSERC) [RGPIN-2024-06637].

\bibliography{references}

\newpage

\end{document}